# Agent Economics: An Entropy-Controlled Pluralistic Alignment Framework for Preventing Artificial Hivemind in Autonomous Agents

Dr. Cheonsu Jeong*

*AX Center, SAMSUNG SDS, Seoul, South Korea*
*paripal@korea.ac.kr*

This study proposes the Behavioral Protocol Framework (BPF), an entropy-controlled pluralistic alignment framework designed to address two critical challenges in autonomous agent economies: the hivemind effect arising from excessive strategic convergence among agents and the lack of transparency in autonomous decision-making processes. The proposed BPF consists of three core modules: Mentalizing-based Social Intelligence (MbSI) grounded in Theory of Mind (ToM), Pluralistic Alignment (PA), and a Verifiable Execution Kernel (VEK). These modules are organically integrated within a closed-loop architecture that governs the entire lifecycle of agent behavior, from decision-making and execution to verification and feedback. To evaluate the proposed framework, a simulation environment implemented in Python and a Streamlit-based user interface will be developed. Through empirical experimentation, the study aims to examine whether the entropy-control mechanism of the PA module can effectively preserve strategic diversity among agents and mitigate collective convergence, while the VEK module provides a comprehensive and transparent audit trail of the decision-making process. The anticipated results are expected to demonstrate that the proposed framework can simultaneously enhance the stability, efficiency, and trustworthiness of autonomous agent economies. Consequently, this research offers a practical approach for developing robust, transparent, and accountable agent-native economic systems.

## 1. Introduction

Recent advances in Artificial Intelligence (AI) have progressed at an unprecedented pace, driving transformative changes across society and various industrial sectors [1]. These technological developments have significantly accelerated the adoption of Autonomous Agents in a wide range of application domains. In particular, agents powered by Large Language Models (LLMs) possess sophisticated capabilities for complex decision-making and interactive reasoning, enabling them to function as key participants in autonomous economic systems such as digital marketplaces, financial transactions, and supply chain management environments [2]. These agents have the potential to maximize the efficiency of economic activities by analyzing data, negotiating, and executing transactions without human intervention. The alignment problem—ensuring advanced AI systems act in accordance with human values—represents one of the most pressing philosophical and technical challenges of our time [3].

* Corresponding author: csu.jeong@samsung.com

However, the proliferation of autonomous agents brings about new challenges. In particular, the phenomenon of groupthink, where agents sharing identical or similar algorithms and learning data make similar decisions in certain situations, known as the hivemind effect, can undermine the overall stability of the system [4]. For example, in financial markets, if numerous AI agents follow the same investment strategy, there is a risk of triggering rapid market volatility or even causing a collapse. Such groupthink reduces market diversity and is a major cause of increasing unpredictable systemic risks [5].

Moreover, the opacity of the agent's decision-making process is another significant issue. Existing AI systems tend to primarily record only the final outcomes, making it difficult to provide clear evidence of the reasons or processes behind specific decisions made by the agent. This poses serious limitations for IT auditing, legal accountability, and ensuring system reliability [8]. To guarantee the reliability of autonomous economic systems, every decision-making step of the agent must be transparently recorded and verifiable.

This study aims to address two key issues in the aforementioned autonomous agent-based economic systems: market instability caused by groupthink and the lack of transparency in the decision-making process. Previous studies have focused on improving the internal decision-making logic of agents or recording transaction outcomes, but have failed to comprehensively resolve the problems of increasing strategic similarity among agents and the inability to trace negotiation processes [2]. In particular, in the case of LLM-based agents, the complexity and nonlinearity of the models make it even more challenging to trace and explain the decision-making process.

The absence of a mechanism to measure and control the diversity of agent strategies in real-time, utilizing the concept of information entropy, poses a significant obstacle to effectively preventing groupthink. Furthermore, the lack of a robust framework to document and post-verify the reasoning paths of agents makes it challenging to ensure fairness and accountability in autonomous economic systems [9].

The purpose of this study is to propose an entropy control-based pluralistic alignment system and method framework, referred to as the Behavioral Protocol Framework (BPF), which aims to regulate interactions among autonomous agents, prevent groupthink, and ensure transparency and reliability in decision-making. Specifically, the following objectives are pursued:

1. **Advanced Mentalizing Logic:** Through a Mentalizing-based Social Intelligence (MbSI) module based on LLMs, it automatically derives advanced negotiation strategies by inferring the hidden intentions, constraints, and preferences of the opposing agent.
2. **Ensuring Strategic Diversity:** By employing the Pluralistic Alignment (PA) algorithm, which utilizes information entropy, we prevent strategy convergence and groupthink among agents, thereby enforcing strategic diversity within the market to simultaneously achieve competitiveness and stability.

3. **Decision Integrity Verification:** By recording and making verifiable the entire decision-making process of an agent through the Verifiable Execution Kernel (VEK) based on a hash chain, IT auditing and legal defensibility are ensured.
4. **Establishing a Closed-Loop Control System:** This study provides a framework for implementing a closed-loop autonomous economic system that integrates agent decision-making, execution, verification, and feedback while enabling coordination with external regulatory and supervisory mechanisms. The proposed architecture facilitates comprehensive governance of agent behavior throughout the entire operational lifecycle.

This research makes the following key contributions:

- **Proposed Integrated Framework:** We propose a BPF that organically integrates the MbSI, PA, and VEK modules, presenting a novel approach to holistically address the core issues of autonomous agent economic systems.
- **Entropy-Based Diversity Control:** This mechanism applies the concept of information entropy to controlling the diversity of agent strategies, thereby proactively preventing groupthink and maintaining a healthy market ecosystem.
- **Verifiable Decision-Making Pathways:** By transparently and immutably recording the chain of reasoning of an agent through a hash chain-based VEK, it dramatically enhances the reliability and accountability of AI systems.
- **Empirical Simulation and UI Implementation:** The core logic of the proposed framework will be implemented in Python. To empirically evaluate the practical effectiveness of the theoretical framework, a simulation environment will be developed using a Streamlit-based user interface. This environment will enable interactive experimentation, parameter configuration, and real-time visualization of agent behaviors and system outcomes.

The structure of this research is as follows. Chapter 2 reviews related prior studies to this research. Chapter 3 presents the detailed design of the proposed BPF and explains the operational principles of each module. Chapter 4 describes the implemented simulation environment and experimental scenarios. Finally, Chapter 5 summarizes the conclusions of this study and suggests directions for future research.

## 2. Related Work

### 2.1. *Autonomous Agents in Digital Marketplaces*

Automated trading and negotiation systems have evolved from early rule-based agents [10] to reinforcement learning-based agents [11], and more recently to architectures powered by LLMs [12]. Recent studies have demonstrated that LLM-based agents are capable of conducting multi-round bilateral negotiations with performance approaching that of human participants in structured negotiation tasks. Despite these advances, most existing evaluations focus primarily on the performance of individual agents, while paying limited attention to market-level characteristics such as price discovery efficiency, the Herfindahl–Hirschman Index (HHI), and bid–ask spread distributions under correlated strategic behavior.

In parallel, growing interest in understanding the broader economic implications of autonomous agents has led to the emergence of Agent-Based Computational Economics (ACE) [13]. ACE employs multi-agent systems (MAS) to simulate complex economic phenomena, enabling researchers to investigate emergent behaviors, market dynamics, and policy interventions in ways that are difficult to capture using traditional equilibrium-based models. This approach has proven particularly valuable for analyzing issues such as supply chain resilience, financial market stability, and the evolution of trading strategies. The transition of agent-based modeling from a primarily theoretical framework to a practical simulation methodology represents a significant milestone in the mainstream adoption of computational approaches within economics. Furthermore, the ability of autonomous agents to perform automated negotiation [14] and matching in electronic marketplaces has long been an active area of research, laying the foundation for today's autonomous agent-driven market ecosystems.

### 2.2. *Autonomous Agents and Theory of Mind (ToM)*

Autonomous agents refer to software or hardware systems that interact with their environment and make independent decisions to achieve specific goals. Recent advancements in LLM have significantly enhanced the cognitive and interactive capabilities of these agents, enabling them to operate in increasingly complex social environments [7].

In order for these agents to effectively cooperate or compete with humans or other agents, the ability to infer the intentions, beliefs, knowledge, emotions, and other mental states of others—known as Theory of Mind (ToM)—is essential [15]. ToM is a core component of human social intelligence, referring to the cognitive ability to understand and predict the mental states of other entities. In the field of artificial intelligence, research is actively underway to endow agents with ToM capabilities using LLM [16, 17]. For example, studies have been introduced in which LLM-based agents infer minimum acceptance prices or cooperation indices in negotiation scenarios based on the counterpart's past transaction history and current offers [6, 16]. These studies contribute to enabling agents to move beyond simply acting according to rules and to develop advanced strategies that reflect the hidden constraints and tendencies of their counterparts.

### 2.3. *AI Alignment and the Hivemind Effect*

AI alignment is a research field that ensures the goals and behaviors of artificial intelligence systems are in accordance with human values, intentions, and objectives [18]. Early AI alignment research primarily focused on aligning the goals of individual agents with human values; however, in complex systems where multiple agents interact, the importance of pluralistic alignment is being emphasized [19, 18]. Pluralistic alignment aims to coordinate various agents within a system to pursue their respective goals while maintaining the overall stability and diversity of the system.

However, the phenomenon of groupthink, where numerous agents sharing identical or similar algorithms, data, and reward functions make similar decisions in specific situations (referred to as the Hivemind Effect), presents a significant challenge for AI alignment [4], [5]. Groupthink can undermine market efficiency and lead to unpredictable systemic risks. For instance, in financial markets, if a multitude of AI trading agents respond to the same market signal by taking short positions, it could trigger a sharp market decline. To prevent such occurrences, it is necessary to maintain strategic diversity among agents and implement mechanisms that inhibit convergence toward specific strategies [20].

### 2.4. *Information Theory and Strategic Diversity Control*

Information theory is effectively utilized to measure the degree of uncertainty or diversity within a system. In particular, information entropy serves as a metric to quantify the uncertainty of a probability distribution and can be applied to measure and control the strategic diversity of agent systems [21, 22]. As the strategy distribution of agents converges toward a specific point, entropy decreases, which can be interpreted as a sign of groupthink. Conversely, when the strategy distribution is widely dispersed, entropy increases, indicating healthy diversity within the system.

Recent studies have proposed methods to enhance the diversity of multi-agent systems by utilizing information entropy [21, 23, 24]. For example, when an agent's behavior exhibits entropy below a certain threshold, diversity is enforced by intentionally introducing perturbations to its strategy or encouraging the exploration of new strategies. This approach helps prevent agent systems from converging to a single optimal solution and contributes to improving adaptability to changing environments.

### 2.5. *Auditable AI and Verifiable Execution*

To achieve true autonomy and enhance trustworthiness in LLM-based systems, the incorporation of self-evolving capabilities and reliability-aware self-healing mechanisms is essential for building a more robust and dependable AI ecosystem [25, 26]. In this context, ensuring transparency and trustworthiness throughout the decision-making process of autonomous agent systems is critical for establishing accountability in verifiable execution frameworks. In particular, when AI agents make economically or socially significant decisions, it is necessary to maintain clear, tamper-resistant, and auditable records that document how such decisions were generated and executed [8]. Verifiable Execution technology is a methodology that utilizes blockchain technology to transparently record and verify the decision-making processes and execution outcomes of agents, thereby meeting these requirements [9, 27].

The hash chain structure of blockchain is a core mechanism that ensures the integrity and immutability of data [28]. Inputs, inference processes, and outputs at each decision-making stage are structured as records, and a new hash value is generated by combining the current record with the hash value of the previous record, thereby linking the chain. As a result, if a record at any point is altered, all subsequent hash values will change, enabling immediate detection of tampering [29, 30]. This technology plays a crucial role in documenting the 'reasoning path' of AI agents, thereby facilitating post-audit and providing legal accountability [8, 30]. Recently, concepts such as the 'Verifiable AI Stack' have emerged, accelerating effrts to build trustworthy AI systems through the integration of AI and blockchain technologies [9].

## 3. Design of Behavioral Protocol Framework(BPF)

This chapter presents the detailed design of the Behavioral Protocol Framework (BPF), an entropy control-based pluralistic alignment system and method for preventing groupthink in agents. The BPF aims to regulate interactions among autonomous agents, ensure strategic diversity, and guarantee transparency and reliability in decision-making.

### 3.1. *System Architecture*

The proposed BPF has a hierarchical structure consisting of an external regulatory and supervisory system (ES), the BPF itself, and an autonomous digital marketplace. These components are organically interconnected to form a closed-loop control mechanism. Figure 1 illustrates the conceptual architecture of this system.

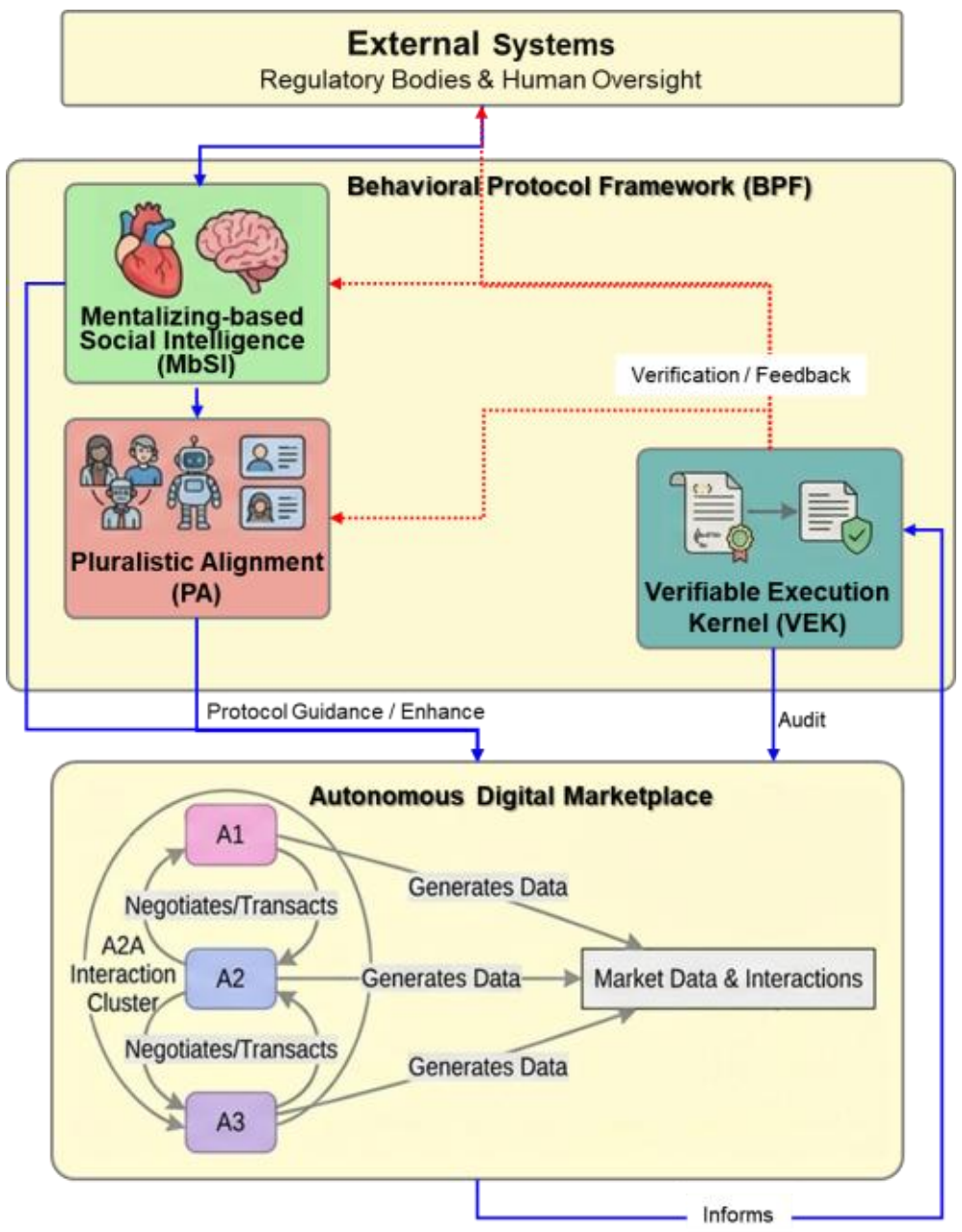

Fig. 1. Conceptual Architecture of the Behavioral Protocol Framework (BPF)

Transaction data generated in autonomous digital marketplaces is delivered to the BPF for analysis and verification. The BPF transmits these results to external regulatory and supervisory systems for audit purposes and policy implementation. Based on the verification outcomes, the external regulatory and supervisory systems develop new policies and constraints, which are then relayed back to the BPF. These updates are integrated into the agents' decision-making processes, enabling dynamic control of the entire system. The BPF consists of three core modules: MbSI, PA, and VEK. These modules are organically interconnected to oversee the entire process, from agents' decision-making and execution to verification and feedback.

## 3.2. *Mentalizing-based Social Intelligence (MbSI)*

The MbSI module plays a role in enhancing negotiation strategies by inferring the hidden intentions, beliefs, and constraints of the opposing agent. This is achieved by utilizing an LLM-based ToM module [7, 16].

### *3.2.1. Processing Procedure*

- The input values include historical transaction data (proposed price, final executed price, number of negotiation rounds, whether there was an intermediate withdrawal, discount or surcharge ratio, etc.) and the current proposal data from the counterpart agent (price, quantity, transaction terms, timestamp, etc.).
- The system normalizes past transaction history and current proposal data and converts them into a state vector *x* = *(price, quantity, negotiation stage, discount rate, churn flag, ...)* representing each transaction point. This state vector is provided as input to the LLM-based ToM module, where the LLM is injected with a structured vector along with a natural language description in the form of a prompt. Through internal reasoning, the LLM generates the intention parameters of the opposing agent $(p_{min}, p_{target}, \Delta p, c)$.

### *3.2.2. Intention Parameter Estimation*

The LLM-based ToM module approximates the conditional probability distribution to estimate the following intent parameters:

- $p_{min}$ : Minimum Acceptable Price of the Counterpart Agent
- $p_{target}$ : Target Price of the Counterparty Agent
- $\Delta p$ : The Concession Range of the Opposing Agent
- $c$ : Cooperation Score of the Counterparty Agent

Based on these intent parameters, the MbSI module determines the corresponding negotiation strategy parameter vector. For example, if the value is low, rules can be applied to reduce the maximum number of negotiation rounds and to set the initial offer price more conservatively. The derived strategy parameters are recorded in the internal state storage of the BPF and utilized in subsequent stages.

### 3.3. *Pluralistic Alignment (PA)*

The PA module enforces strategic diversity in the market, preventing strategy convergence and groupthink among agents, while simultaneously ensuring competitiveness and stability [18, 19]. This is achieved by measuring and controlling the diversity of agent strategies based on information entropy [21, 23].

#### *3.3.1. Processing Procedure*

- Use the set of strategy parameters $S_j$ of other agents active at the same time as the strategy parameter $S_1$ generated in MbSI as input.
- To calculate similarity, embed each strategy parameter $S_j$ into a fixed-length vector. For example, a vector composed of price-related parameters, the number of negotiation rounds, and the possible concession range can be used. Then, use a cosine similarity or inner product-based function to calculate the similarity $sim(S_1, S_j)$ between the target agent strategy $S_1$ and the other agent strategy $S_j$.

#### *3.3.2. Calculation of Information Entropy and Strategy Adjustment*

Based on the similarity values, strategy types are divided into several clusters, and the strategy distribution *P(k)* is defined through the proportion of strategies belonging to each cluster. Using this strategy distribution P(k), the information entropy *H* is calculated. Entropy *H* is defined as follows:

$$H = -\sum_{k} P(k) \log P(k)$$

Here, *P(k)* represents the proportion of agents belonging to the k-th strategy cluster. A state where the entropy H is below a preset threshold ($H^{thd}$) is defined as a strategy convergence state, and is considered a hivemind state. When it is judged to be in a hivemind state, the PA module applies a strategic perturbation to $S_1$ to generate the adjusted strategy parameters $S_2$. This adjustment process actively maintains strategic diversity by making limited changes to key control variables such as price-related parameters, concession range, and the number of negotiation rounds. The adjusted strategy parameters $S_2$ are then passed on to the subsequent negotiation stage.

### 3.4. *Verifiable Execution Kernel (VEK)*

The VEK module is responsible for transparently and immutably recording and verifying the entire decision-making process of an agent [8, 30]. This is implemented by utilizing the hash chain structure of the blockchain [9, 28].

#### *3.4.1. Recording and Processing*

- The input values include the initial state of each negotiation round (price, terms), LLM inference summary text, output proposal price, and other summarized inference stage information from the negotiation process, as well as the final agreement data.
- For each negotiation round t, the input state, LLM inference summary, output proposed prices, etc., are structured into a single record $R_t$. The serialized byte sequence of this record $R_t$ is combined with the hash value of the previous round $HC_{t-}$ $_1$ to calculate a new hash value $HC_t$ using a hash function $h(\cdot)$ (e.g., SHA-256). The hash chain is generated in the following manner:

$$HC_t = h(R_t \parallel\parallel HC_{t-1})$$

Here, $HC_0$ is the initial hash value (e.g., a string filled with 0). For the entire negotiation rounds, generate $R_t$, $HC_t$ pairs, and configure the hash value $HC_t$ of the last round and the final agreement data as the block header, and store it in the ledger (log database or blockchain network).

#### ***3.4.2.*** *Validation*

In the event of a post-audit or dispute, the hash chain can be recalculated in the same manner from the stored record set $R_t$ and compared with $HC_t$ stored in the ledger to verify whether the entire execution flow has been tampered with. Through this process, VEK provides a complete and verifiable record of AI agent execution, referred to as the "Right to History," to all stakeholders by storing the entire decision-making process in a traceable format, rather than merely logging the final results [8].

### **3.5.** ***Five-Stage Processing Pipeline for Execution***

The BPF pipeline processes each inter-agent transaction through five sequential stages. Table 1 outlines the inputs, processes, and outputs for each stage.

Table 1. Summary of the Five-Stage BPF Pipeline

| **Phase** | **Module** | **Process Overview** | **Output** |
|---|---|---|---|
| 1 | MbSI | Construction of state vectors; inference of theory of mind in large language models; extraction of intent parameters; mapping of strategies | Strategic Vector $S_1$ |
| 2 | PA | Strategic embedding; cosine similarity clustering; entropy calculation; conditional perturbation | Revised Strategy $S_2$ |
| 3 | Negotiation | Iterative bilateral offer-counteroffer under $S_2$ parameters; convergence or termination | Contract Object $C$ |
| 4 | VEK | Round-by-round record structuring; hash-chain formation; ledger commitment | Immutable audit *log* $\{R_t, H_t\}$ |
| 5 | Execution & Feedback | Asset transfer and settlement; feedback incorporation into MbSI/PA databases; ES policy revision | Completed transaction and learning data |

#### ***3.5.1.*** *Stage 1: Intent Analysis via MbSI*

Each transaction begins with an intent analysis cycle. The agent's local BPF instance retrieves the counterpart's historical transaction records from the shared ledger (where available under privacy constraints), constructs normalized state vectors, and submits the prompt package to the ToM inference module. The output strategy vector $S_1$ captures a best-response to the inferred counterpart model and is stored in the BPF's internal state buffer.

#### 3.5.2. Stage 2: Pluralistic Alignment Filtering

Before strategy $S_1$ is committed, the PA module samples the current population strategy distribution from the ADM broker. If $H(P) \geq H^{thr}$, $S_1$ is passed through unchanged. If $H(P) < H^{thr}$, the perturbation subroutine is applied, producing $S_2 \neq S_1$. The BPF logs the perturbation event, including the pre- and post-perturbation entropy values and the perturbation delta, for ES review.

#### 3.5.3. Stage 3: Autonomous Negotiation

The negotiating agent executes multi-round bilateral negotiation under the parameters encoded in $S_2$. Each round follows a standard alternating-offers protocol [31]: the agent transmits an offer; the counterpart responds with acceptance or a counteroffer; the agent updates its offer according to the concession schedule defined in $S_2$. Negotiation terminates upon mutual acceptance or exhaustion of $n^{max}$ rounds. The agreed contract object $C$ is passed to VEK.

#### 3.5.4. Stage 4: VEK Recording and Verification

For each round $t$, the VEK constructs and chains record $R_t$ as described in Section 3.4. Upon negotiation conclusion, the terminal hash $H^T$ and contract $C$ are submitted to the ledger. The VEK module returns a commitment receipt containing the ledger entry identifier, enabling subsequent verification without exposing proprietary negotiation details.

#### 3.5.5. Stage 5: Execution and Closed-Loop Feedback

The ADM execution engine settles the contract, transferring assets and recording the outcome (settlement price, latency, success status). Execution outcomes are appended to the MbSI training database for future ToM inference improvement and to the PA module's strategy distribution cache. Periodically, ES ingests batch VEK logs, computes market-level statistics (mean strategy entropy, HHI, collusion risk scores), and issues updated constraint vectors to all BPF instances, closing the regulatory feedback loop.

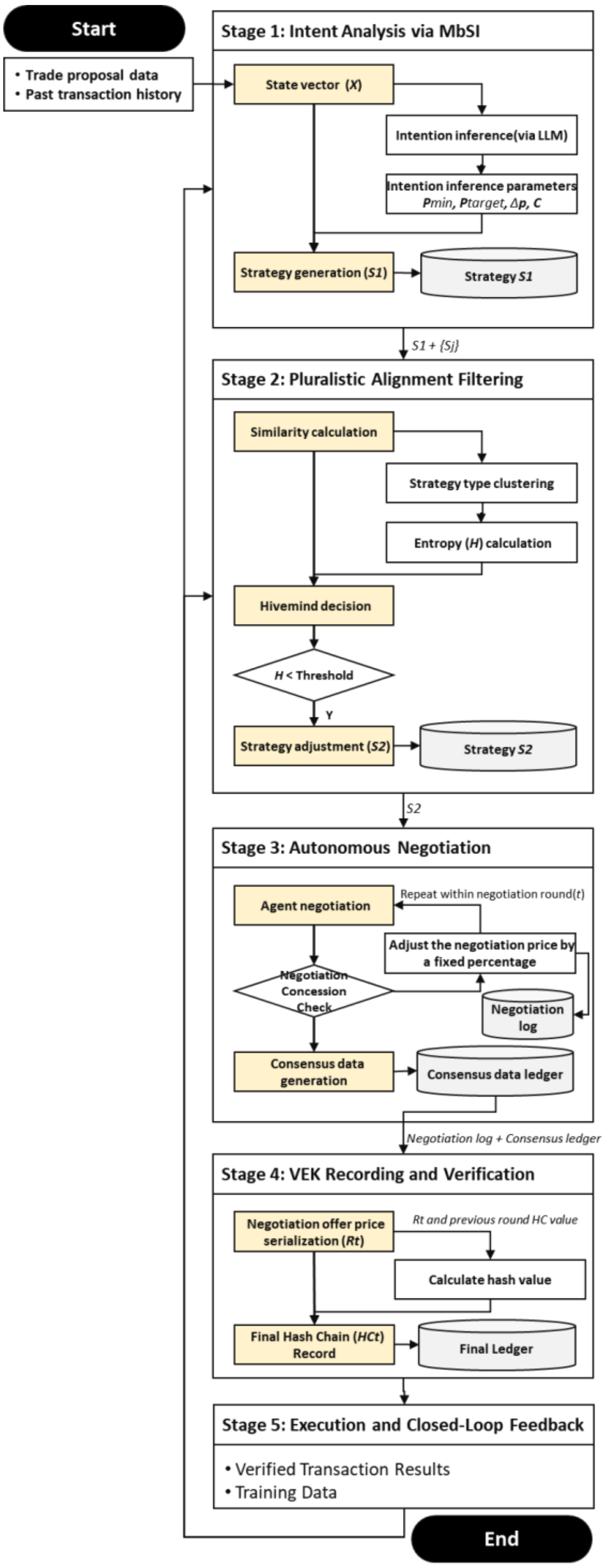

Fig. 2. Hierarchical Architecture Flow

# 4. Empirical experiments

This chapter presents the simulation environment and experimental setup designed to demonstrate the operation of the core mechanisms of the proposed BPF, namely MbSI, PA, and VEK modules. Particular emphasis is placed on evaluating the effectiveness of entropy-based pluralistic alignment in preventing artificial hivemind behavior among autonomous agents, as well as assessing the verifiability and transparency of the decision-making process.

## 4.1. *Establishment of the experimental environment*

To conduct the empirical evaluation, a Python-based simulation environment will be developed. The core BPF logic, including the MbSI, PA, and VEK modules, will be implemented within the simulation framework, while the UI will be developed using Streamlit. Through the Streamlit-based UI, users will be able to configure key experimental parameters, including the initial market price, the entropy threshold of the PA module, and the number of neighboring agents. Users will also be able to execute simulations and visualize results in real time. The sidebar interface will allow users to adjust the initial market conditions and diversity control parameters. In particular, a "Hivemind Scenario Simulation" option will be provided to configure neighboring agents with highly similar strategies, thereby creating conditions that increase the likelihood of collective strategic convergence. Upon execution of the simulation, the MbSI, PA, and VEK modules will operate sequentially. Key performance indicators, including the final agreement price, strategy entropy, and VEK verification results, will be presented through an interactive dashboard. In addition, negotiation-round price trajectories and agent strategy-space distributions will be visualized to facilitate intuitive understanding of changes in strategic diversity over time. The VEK hash-chain log will be displayed in tabular form to provide transparency and traceability throughout the decision-making process.

## 4.2. *Experimental Scenario*

The effectiveness of BPF was validated through two main scenarios.

### *4.2.1. Scenario 1: Changes in Strategic Diversity According to Entropy Control*

This scenario investigates the impact of the entropy-control mechanism of the PA module on the strategic diversity of autonomous agents. To emulate a potential hivemind condition, the “Hivemind Scenario Simulation” option will be activated, causing neighboring agents to adopt highly similar negotiation strategies. Simulations will then be conducted under varying entropy-threshold settings of the PA module. The objective is to evaluate how entropy-based interventions influence the distribution of agent strategies and maintain strategic diversity within the marketplace.

### *4.2.2. Scenario 2: Transparency and Verifiability of the Decision-Making Process through VEK*

This scenario examines how the VEK module records and verifies agent negotiation and decision-making processes. Following the execution of the simulation, the VEK hash-chain logs will be analyzed to verify that critical information—including input states, reasoning summaries, and generated offer prices for each negotiation round—has been immutably recorded. This experiment aims to demonstrate the transparency, traceability, and verifiability of the decision-making process provided by the VEK mechanism.

## 5. Conclusion and Discussion

### 5.1. *Conclusion*

This study proposes an Entropy-Controlled Pluralistic Alignment Framework, referred to as BPF, to address two critical challenges in autonomous agent economies: the emergence of artificial hivemind behavior and the lack of transparency in autonomous decision-making processes.

The proposed BPF consists of three integrated modules: MbSI, PA, and the VEK. These modules collectively establish a closed-loop architecture that governs the entire lifecycle of agent behavior, from decision-making and execution to verification and feedback.

The MbSI module utilizes ToM capabilities supported by large language models to infer the intentions of counterpart agents, thereby enabling the formulation of more sophisticated negotiation strategies. The PA module continuously measures the strategic diversity of the agent population through information entropy. When entropy falls below a predefined threshold, the module intentionally perturbs agent strategies to mitigate artificial hivemind effects and maintain a healthy level of market diversity. Finally, the VEK module employs a blockchain-inspired hash-chain structure to transparently and immutably record all stages of agent reasoning and decision-making, enabling the measurement and verification of AI-system trustworthiness and accountability.

The core mechanisms of the proposed BPF will be evaluated through a Python-based simulation environment and a Streamlit-based user interface. The experimental results are expected to demonstrate that the entropy-control mechanism of the PA module effectively preserves strategic diversity and mitigates collective convergence, while the VEK module provides a comprehensive audit trail for autonomous decision-making processes.

Overall, the anticipated findings are expected to provide practical evidence that the proposed framework can simultaneously enhance the stability, efficiency, transparency, and trustworthiness of autonomous agent economies, thereby contributing to the development of robust and responsible agent-native marketplaces.

### 5.2. *Discussion and Future Research Directions*

The BPF proposed in this study contributes to addressing key issues in autonomous agent-based economic systems, but it also has several limitations and suggests directions for future research.

First, the LLM-based ToM inference of the MbSI module was implemented in a simplified manner within the simulation. In real-world environments, it is necessary to optimize the inference performance and efficiency of the LLM, and to enhance the module's ability to learn various types of agent behavioral patterns for more sophisticated intention inference. In particular, in-depth research is required on the ability to predict and respond to the behaviors of non-cooperative or malicious agents.

Second, the entropy calculation and strategy adjustment mechanism of the PA module currently employs fixed threshold values. Research is needed on adaptive entropy control algorithms that can dynamically adjust threshold values according to market conditions or agent characteristics, or that can modulate the responsiveness to entropy reduction. Furthermore, additional analysis is required regarding the optimal intensity and methods of strategic perturbations.

Third, while the VEK module provides hash chain-based recording and verification, an in-depth analysis of its performance and scalability in real large-scale distributed environments is necessary. Optimization strategies should be investigated, taking into account practical implementation issues such as transaction costs, processing speed, and consensus mechanisms that may arise when directly integrated with a blockchain network.

Finally, the simulation of this study was conducted based on a limited number of agents and simplified negotiation scenarios. Future research should establish large-scale simulation environments involving hundreds or thousands of agents and apply actual financial market data or complex supply chain scenarios to verify the robustness and scalability of BPF. Additionally, studies could be conducted to validate the effectiveness of BPF across various industries and to provide concrete guidelines for its implementation as a practical service.

Future research is expected to further develop BPF into a more robust and practical foundational technology for autonomous agent-based economic systems.

**Acknowledgments**

This study is an expanded research that includes empirical components based on the content of a South Korean patent application (Application Number: 10-2026-0092857).

**References**

[1] Jeong, C., Lee, S., Jeong, S., & Kim, S. (2026). A Study on the Framework for Evaluating the Ethics and Trustworthiness of Generative AI: A Case of Generative AI Chatbot Services. *Artificial Intelligence and Applications*. https://orcid.org/0000-0003-0751-3987

[2] Deture, S. (2025). Policy entropy, learning, and alignment (Or Maybe Your LLM Needs Therapy). https://sdeture.substack.com/p/policy-entropy-learning-and-alignment

[3] Spizzirri, A. (2025). Investigating syntropic frameworks in AI alignment: A philosophical viewpoint. *arXiv*. https://arxiv.org/html/2512.03048v1

[4] Emergent Mind (2025). Artificial hivemind effect. Emergent Mind. https://www.emergentmind.com/topics/artificial-hivemind-effect

[5] Nehalmr. (2025). Analysis of the hivemind effect in AI agents: A comprehensive teaching guide. Medium. https://nehalmr.medium.com/analysis-of-the-hivemind-effect-in-ai-agents-a-comprehensive-teaching-guide-7a6abf8783ed

[6] J. Wu, Y. (2025). An agent-based emotional persuasion model driven by large language models. Engineering Applications of Artificial Intelligence. ScienceDirect. https://www.sciencedirect.com/science/article/abs/pii/S0952197625005676

[7] Hwang, E., Yin, Y., Carenini, G., West, P., & Shwartz, V. (2025). Infusing Theory of Mind into Socially Intelligent LLM Agents. *arXiv preprint arXiv:2509.22887*.

[8] Zhang, J. (2026). Right to History: A Sovereignty Kernel for Verifiable AI Agent Execution. *arXiv preprint arXiv:2602.20214.*

[9] Chainlink. (2026). The verifiable AI stack explained. Chainlink. https://chain.link/article/verifiable-ai-stack

[10] Arthur, W. B. (1994). Inductive reasoning and bounded rationality. *American Economic Review*, 84(2), 406–411.

[11] Nevmyvaka, Y., Feng, Y., & Kearns, M. (2006). Reinforcement learning for optimized trade execution. *In Proceedings of the 23rd International Conference on Machine Learning (ICML 2006)*.

[12] Park, J. S., O'Brien, J. C., Cai, C. J., Morris, M. R., Liang, P., & Bernstein, M. S. (2023). Generative agents: Interactive simulacra of human behavior. *In Proceedings of the 36th Annual ACM Symposium on User Interface Software and Technology (UIST '23). ACM*.

[13] Tesfatsion L (2006) Agent-based computational economics: A constructive approach to economic theory. *In: Handbook of computational economics*, vol 2. Elsevier, pp 831-880

[14] Jennings NR, et al. (2001) Automated negotiation: prospects, methods and challenges. *International Journal of Group Decision and Negotiation* 10(2):199-215

[15] Kanishk Patel. (2026). What is theory of mind for AI agents? Substack. https://learnagentic.substack.com/p/what-is-theory-of-mind-for-ai-agents

[16] Ioguntol. (2024). Investigating the theory of mind in large language models via multimodal negotiation. *ACM*. https://dl.acm.org/doi/10.1145/3652988.3673960

[17] Carnegie Mellon University. (2025). Theory of mind in multi-agent systems (Doctoral dissertation). https://ml.cmu.edu/research/phd-dissertation-pdfs/ioguntol_phd_mld_2025.pdf

[18] W. Zeng, H., Zhu, C., Qin, H., Wu, Y., Cheng, & Zhang, S. (2025). Multi-level value alignment in agentic AI systems: Survey and perspectives. *arXiv*. https://arxiv.org/abs/2506.09656

[19] Taylor Sorensen. (2024). Pluralistic alignment: A roadmap, recent work, and open problems [Video]. YouTube. https://www.youtube.com/watch?v=1F0iaivYdvI

[20] N. Mamie, & S. R. Xi. (2025). The society of HiveMind: Multi-agent optimization of foundation model swarms to unlock the potential of collective intelligence. *In Proceedings published by Springer*. https://link.springer.com/chapter/10.1007/978-981-95-0982-9_20

[21] De Paola, V., Zamboni, R., Mutti, M., & Restelli, M. (2025). Enhancing diversity in parallel agents: A maximum state entropy exploration story. *arXiv preprint arXiv:2505.01336.*

[22] Rother, D., Herbert, F., Kalter, F., Koert, D., Pajarinen, J., Peters, J., & Weisswange, T. H. (2025). Entropy-based blending of policies for multi-agent coexistence. *Autonomous Agents and Multi-Agent Systems*, 39(1), 27.

[23] Li, T., & Zhu, K. (2025, May). Self-Supervised Multi-Agent Diversity with Nonparametric Entropy Maximization. *In Proceedings of the 24th International Conference on Autonomous Agents and Multiagent Systems* (pp. 1291-1299).

[24] Yao, J., Cheng, R., Wu, X., Wu, J., & Tan, K. C. (2026). Diversity-aware policy optimization for large language model reasoning. *Advances in Neural Information Processing Systems*, 38, 94801-94826.

[25] Jeong, C., Kim, Y., & Shin, K. (2026). Self-Evolving Multi-Agent Framework (SEMAF): Continuous Learning and Adaptive Collaboration for Robust LLM-Based Systems. *The Korea Journal of BigData*, 11(1), 1-15.

[26] Jeong, C., & Shin, Y. (2026). A Self-Healing Framework for Reliable LLM-Based Autonomous Agents. arXiv preprint arXiv:2605.06737.

[27] VerticalServe. (2026). Blockchain — How "verifiable AI/agents" will unlock the next generation of dApps. Medium. https://verticalserve.medium.com/blockchain-how-verifiable-ai-agents-will-unlock-the-next-generation-of-dapps-95df3fe3d5f6

[28] Kanishk Patel (2026). What is Theory of Mind for AI Agents? https://learnagentic.substack.com/p/what-is-theory-of-mind-for-ai-agents

[29] Arnold Hayes. (2025). Blockchain and AI: Building on each other for verifiable data. LinkedIn. https://www.linkedin.com/posts/arnold-hayes_blockchain-artificialintelligence-web3-activity-7442920497731952640-emv0

[30] Zhang, J. (2026). Right to History: A Sovereignty Kernel for Verifiable AI Agent Execution. *arXiv preprint arXiv:2602.20214*.

[31] Rubinstein, A. (1982). Perfect equilibrium in a bargaining model. *Econometrica*, 50(1), 97–109.